\documentclass{article} 
\usepackage{iclr2015,times}
\usepackage{hyperref}
\usepackage{url}
\usepackage[british,UKenglish,USenglish,english,american]{babel}
\usepackage{amsmath}
\usepackage{subfig}
\usepackage{graphicx}
\usepackage{comment}
\usepackage{multirow}

\usepackage{mathtools}
\usepackage[toc,page]{appendix}

\def\E{{\rm E}}
\def\D{{\rm D}}
\def\G{{\rm G}}

\title{Generative Modeling of Convolutional Neural Networks}

\author{
Jifeng Dai\\
Microsoft Research\\
\texttt{jifdai@microsoft.com}
\And
Yang Lu and Ying Nian Wu\\
University of California, Los Angeles\\
\texttt{$\{$yanglv, ywu$\}$@stat.ucla.edu}
}

\iclrfinalcopy 

\iclrconference 

\begin{document}

\maketitle

\begin{abstract}
This paper investigates generative modeling of the convolutional neural networks (CNNs). The main contributions include: (1) We construct a generative model for CNNs in the form of exponential tilting of a reference distribution. (2) We propose a generative gradient for pre-training CNNs by a non-parametric importance sampling scheme, which is fundamentally different from the commonly used discriminative gradient, and yet has the same computational architecture and cost as the latter. (3) We propose a generative visualization method for the CNNs by sampling from an explicit parametric image distribution. The proposed visualization method can directly draw synthetic samples for any given node in a trained CNN by the Hamiltonian Monte Carlo (HMC) algorithm, without resorting to any extra hold-out images. Experiments on the ImageNet benchmark show that the proposed generative gradient pre-training helps improve the performances of CNNs, and the proposed generative visualization method generates meaningful and varied samples of synthetic images from a large and deep CNN.
\end{abstract}

\section{Introduction}

Recent years have witnessed the triumphant return of the feedforward neural networks, especially the convolutional neural networks (CNNs) \citep{lecun1989backpropagation,krizhevsky2012imagenet,girshick2014rich}. 
Despite the successes of the discriminative learning of CNNs, the generative aspect of CNNs has not been thoroughly investigated. But it can be very useful for the following reasons: (1) The generative pre-training has the potential to lead the network to a better local optimum; (2) Samples can be drawn from the generative model to reveal the knowledge learned by the CNN. Although many generative models and learning algorithms have been proposed \citep{hinton2006fast,hinton2006unsupervised,rifai2011contractive,salakhutdinov2009deep}, most of them  have not been applied to learning large and deep CNNs.

In this paper, we study the generative modeling of the CNNs. We start from defining probability distributions of images given the underlying object categories or class labels, such that the CNN with a final logistic regression layer serves as the corresponding conditional distribution of the class labels given the images. These distributions are in the form of exponential tilting of a reference distribution, i.e., exponential family models or energy-based models relative to a reference distribution.

With such a generative model, we proceed to study it along two related themes, which differ in how to handle the reference distribution or the null model. In the first theme, we propose a non-parametric generative gradient for pre-training the CNN, where the CNN is learned by the stochastic gradient algorithm that seeks to minimize the log-likelihood of the generative model. The gradient of the log-likelihood is approximated by the importance sampling method that keeps reweighing the images that are sampled from a non-parametric implicit reference distribution, such as the distribution of all the training images. The generative gradient is fundamentally different from the commonly used discriminative gradient, and yet in batch training, it shares the same computational architecture as well as computational cost as the discriminative gradient. This generative learning scheme can be used in a pre-training stage that is to be followed by the usual discriminative training. The generative log-likelihood provides stronger driving force than the discriminative criteria for stochastic gradient by requiring the learned parameters to explain the images instead of their labels. Experiments on the MNIST \citep{lecun1998gradient} and the ImageNet \citep{deng2009imagenet} classification benchmarks show that this generative pre-training scheme helps improve the performance of CNNs.

The second theme in our study of generative modeling is to assume an explicit parametric form of the reference distribution, such as the Gaussian white noise model, so that we can draw synthetic images from the resulting probability distributions of images. The sampling can be accomplished by the Hamiltonian Monte Carlo (HMC) algorithm \citep{neal2011mcmc}, which iterates between a bottom-up convolution step and a top-down deconvolution step. The proposed visualization method can directly draw samples of synthetic images for any given node in a trained CNN, without resorting to any extra hold-out images.  Experiments show that meaningful and varied synthetic images can be generated for nodes of a large and deep CNN discriminatively trained on ImageNet.

\section{Past work}

The generative model that we study is an energy-based model. Such models include field of experts \citep{roth2009fields}, product of experts \citep{hinton2002training}, Boltzmann machines \citep{hinton2006fast}, model based on neural networks \citep{hinton2006unsupervised}, etc. However, most of these generative models and learning algorithms  have not been applied to learning large and deep CNNs. 

The relationship between the generative models and the discriminative approaches has been extensively studied  \citep{jordan2002discriminative,liang2008asymptotic}. The usefulness of generative pre-training for deep learning has been studied by \citet{erhan2010does} etc. However, this issue has not been thoroughly investigated for CNNs. 


As to visualization, our work is related to \citet{erhan2009visualizing,le2012building,girshick2014rich,zeiler2013visualizing,long2014convnets}. In \citet{girshick2014rich,long2014convnets}, the high-scoring image patches are directly presented. In \citet{zeiler2013visualizing}, a top-down deconvolution process is employed to understand what contents are emphasized in the high-scoring input image patches. In \citet{erhan2009visualizing,le2012building,simonyan2013deep}, images are synthesized by maximizing the response of a given node in the network. In our work, a generative model is formally defined. We sample from the well-defined probability distribution by the HMC algorithm, generating meaningful and varying synthetic images, without resorting to a large collection of hold-out images \citep{girshick2014rich,zeiler2013visualizing,long2014convnets}.

\section{Generative model based on CNN}
\subsection{Probability distributions on images}
Suppose we observe images from many different object categories. Let $x$ be an image from an object category $y$. Consider the following probability distribution on $x$,
\begin{eqnarray}
   p_y(x; w) = \frac{1}{Z_y(w)} \exp\left(f_y(x; w)\right) q(x),  \label{eq:G}
\end{eqnarray}
where  $q(x)$ is a reference distribution common to all the categories, $f_y(x; w)$ is a scoring function for class $y$, $w$ collects the unknown parameters to be learned from the data, and  $Z_y(w) = \E_q[\exp(f_y(x; w))] = \int \exp(f_y(x; w)) q(x) dx$ is the normalizing constant or partition function. The distribution $p_y(x; w)$ is in the form of an exponential tilting of the reference distribution $q(x)$, and can be considered an energy-based model or an exponential family model. In Model (\ref{eq:G}), the reference distribution $q(x)$ may not be unique. If we change $q(x)$ to $q_1(x)$, then we can change $f_y(x; w)$ to $f_y(x; w) - \log [q_1(x)/q(x)]$, which may correspond to a $f_y(x; w_1)$ for a different $w_1$ if the parametrization of $f_y(x, w)$ is flexible enough. We want to choose $q(x)$ so that either $q(x)$ is reasonably close to $p_y(x; w)$ as in our non-parametric generative gradient method, or the resulting $p_y(x; w)$ based on $q(x)$ is easy to sample from as in  our generative visualization method.

For an image $x$, let $y$ be the underlying object category or class label, so that $p(x|y; w) = p_y(x; w)$. Suppose the prior distribution on $y$ is $p(y) = \rho_y$. The posterior distribution of $y$ given $x$ is
\begin{eqnarray}
   p(y |x, w) = \frac{\exp(f_y(x; w) + \alpha_y)}{\sum_y \exp(f_y(x; w) + \alpha_y)}, \label{eq:D}
 \end{eqnarray}
where $\alpha_y = \log \rho_y - \log Z_y(w)$. $p(y|x, w)$ is in the form of a multi-class logistic regression, where $\alpha_y$ can be treated as an intercept parameter to be estimated directly if the model is trained discriminatively. Thus for notational simplicity, we shall assume  that the intercept term $\alpha_y$ is already absorbed into $w$ for the rest of the paper. Note that $f_y(x; w)$ is not unique in (\ref{eq:D}). If we change $f_y(x; w)$ to $f_y(x; w) - g(x)$ for a $g(x)$ that is common to all the categories, we still have the same $p(y|x; w)$. This non-uniqueness corresponds to the non-uniqueness of $q(x)$ in (\ref{eq:G}) mentioned above.

Given a set of labeled data $\{(x_i, y_i)\}$, equations (\ref{eq:G}) and (\ref{eq:D}) suggest two different methods to estimate the parameters $w$. One is to maximize the generative log-likelihood $l_{\G}(w) = \sum_i \log p(x_i|y_i, w)$, which is the same as maximizing the full log-likelihood $\sum_i \log p(x_i, y_i|w)$, where the prior probability of $\rho_y$ can be estimated by class frequency of category $y$. The other is to maximize the discriminative log-likelihood $l_{\D}(w) = \sum_i \log p(y_i|x_i, w)$.
For the discriminative model (\ref{eq:D}), a popular choice of $f_y(x; w)$ is multi-layer perceptron or CNN, with $w$ being the connection weights, and the top-layer is a multi-class logistic regression. This is the choice we adopt throughout this paper.

\subsection{Generative gradient}

The gradient of the discriminative log-likelihood is calculated according to
\begin{eqnarray}
    \frac{\partial}{\partial w} \log p(y_i|x_i, w) = \frac{\partial}{\partial w} f_{y_i}(x_i; w) - \E_{\D} \left[ \frac{\partial}{\partial w} f_y(x_i; w)\right],
 \end{eqnarray}
where $\alpha_y$ is absorbed into $w$ as mentioned above, and the expectation for discriminative gradient is
\begin{eqnarray}
    \E_{\D} \left[ \frac{\partial}{\partial w} f_y(x_i; w)\right]  = \sum_{y}  \frac{\partial}{\partial w} f_y(x_i; w)    \frac{\exp(f_y(x_i; w))}{\sum_{y} \exp(f_y(x_i; w))}.
\end{eqnarray}
The gradient of the generative log-likelihood is calculated according to
\begin{eqnarray}
    \frac{\partial}{\partial w} \log p_{y_i}(x_i; w) =  \frac{\partial}{\partial w}  f_{y_i}(x_i; w) - \E_{\G} \left[ \frac{\partial}{\partial w}  f_{y_i}(x; w) \right],
 \end{eqnarray}
 where the expectation for generative gradient is
 \begin{eqnarray}
     \E_{\G} \left[ \frac{\partial}{\partial w}  f_{y_i}(x; w) \right] = \int \frac{\partial}{\partial w}  f_{y_i}(x; w)  \frac{1}{Z_{y_i}(w)} \exp(f_{y_i}(x; w)) q(x),
\end{eqnarray}
which can be approximated by importance sampling. Specifically, let $\{\tilde{x}_j\}_{j=1}^{m}$ be a set of samples from $q(x)$, for instance, $q(x)$ is the distribution of images from all the categories. Here we do not attempt to model $q(x)$ parametrically, instead, we treat it as an implicit non-parametric distribution.  Then by importance sampling,
 \begin{eqnarray}
     \E_{\G}\left[ \frac{\partial}{\partial w}  f_{y_i}(x; w) \right] \approx \sum_j \frac{\partial}{\partial w}  f_{y_i}(\tilde{x}_j; w)  W_j,
\end{eqnarray}
where the importance weight $W_j \propto \exp(f_{y_i}(\tilde{x}_j; w))$ and is normalized to have sum 1. Namely,
\begin{eqnarray}
      \frac{\partial}{\partial w} \log p_{y_i}(x_i; w)   \approx   \frac{\partial}{\partial w}  f_{y_i}(x_i; w)  - \sum_j \frac{\partial}{\partial w}  f_{y_i}(\tilde{x}_j; w)  \frac{\exp(f_{y_i}(\tilde{x}_j; w))}{\sum_k \exp(f_{y_i}(\tilde{x}_k; w))}.
      \label{eq:G_final}
\end{eqnarray}

The discriminative gradient and the generative gradient differ subtly and yet  fundamentally in calculating  $\E [\partial  f_y(x; w) /\partial w ]$, whose difference from the observed $\partial f_{y_i}(x_i; w)/\partial w$ provides the driving force for updating $w$. In the discriminative gradient, the expectation is with respect to the posterior distribution of the class label $y$ while the image $x_i$ is fixed, whereas in the generative gradient, the expectation is with respect to the distribution of the images $x$ while the class label $y_i$ is fixed. In general, it is easier to adjust the parameters $w$ to predict the class labels than to reproduce the features of the images. So it is expected that the generative gradient provides stronger driving force for updating $w$.

The non-parametric generative gradient can be especially useful in the beginning stage of training or what can be called pre-training, where $w$ is small, so that the current $p_y(x; w)$ for each category $y$ is not  very separated from  $q(x)$, which is the overall distribution of $x$.  In this stage, the importance weights $W_j$ are not very skewed and the effective sample size for importance sampling can be large. So updating $w$ according to the generative gradient can provide useful pre-training with the potential to lead $w$ toward a good local optimum. If the importance weights $W_j$ start to become skewed and the effective sample size starts to dwindle, then this indicates that the categories $p_y(x; w)$ start to separate from $q(x)$ as well as from each other, so we can switch to discriminative training to further separate the categories. 

\subsection{Batch training and generative loss layer}
\label{sec:Batch training}

At first glance, the generative gradient appears computationally expensive due to the need to sample from $q(x)$. In fact, with $q(x)$ being the collection of images from all the categories, we may use each batch of samples as an approximation to $q(x)$ in the batch training mode.

Specifically, let $\{(x_i,y_i)\}_{i=1}^{n}$ be a batch set of training examples, and we seek to maximize $\sum_i \log p_{y_i}(x_i;w)$ via generative gradient. In the calculation of $\partial \log p_{y_i}(x_i;w)/\partial w$, $\{x_j\}_{j=1}^{n}$ can be used as samples from $q(x)$. In this way, the computational cost of the generative gradient is about the same as that of the discriminative gradient.

Moreover, the computation of the generative gradient can be induced to share the same back propagation architecture as the discriminative gradient. Specifically, the calculation of the generative gradient can be decoupled into the calculation at a new generative loss layer and the calculation at lower layers. To be more specific, by replacing $\{\tilde{x}_j\}_{j=1}^{m}$ in (\ref{eq:G_final})  by the batch sample $\{x_j\}_{j=1}^{n}$, we can rewrite (\ref{eq:G_final}) in the following form:
\begin{eqnarray}
    \frac{\partial}{\partial w} \log p_{y_i}(x_i;w) \approx \sum_{y,j} \frac{\partial \log p_{y_i}(x_i;w)}{\partial f_y(x_j;w)} \frac{\partial f_y(x_j;w)}{\partial w},
 \end{eqnarray}
where ${\partial \log p_{y_i}(x_i;w)}/{\partial f_y(x_j;w)}$ is called the generative loss layer (to be defined below, with $f_y(x_j;w)$ being treated here as a variable in the chain rule), while the calculation of ${\partial f_y(x_j;w)}/{\partial w}$ is exactly the same as that in the discriminative gradient. This decoupling brings simplicity to programming.

We use the notation ${\partial \log p_{y_i}(x_i;w)}/{\partial f_y(x_j;w)}$ for the top generative layer mainly to make it conformal to the chain rule calculation. According to (\ref{eq:G_final}),  ${\partial \log p_{y_i}(x_i;w)}/{\partial f_y(x_j;w)}$  is defined  by
 \begin{eqnarray}
 \frac{\partial \log p_{y_i}(x_i;w)}{\partial f_y(x_j;w)} =
\begin{dcases}
0& y \neq y_i ;\\
1-\frac{\exp(f_{y_i}(x_j;w))}{\sum_{k} \exp(f_{y_i}(x_k;w))}& y=y_i, j=i;\\
-\frac{\exp(f_{y_i}(x_j;w))}{\sum_{k} \exp(f_{y_i}(x_k;w))}& y=y_i, j \neq i.
\end{dcases}
 \end{eqnarray}

\subsection{Generative visualization}

 \begin{figure}
	\centering
	\setlength{\tabcolsep}{2.1pt}
	\begin{tabular}{cccc}
		\includegraphics[width=.2\textwidth]{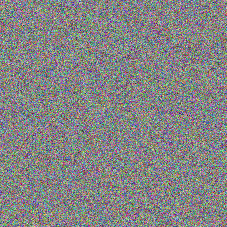} &
		\includegraphics[width=.2\textwidth]{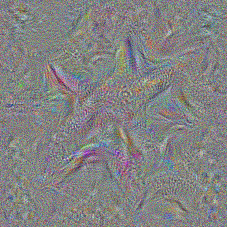} &
		\includegraphics[width=.2\textwidth]{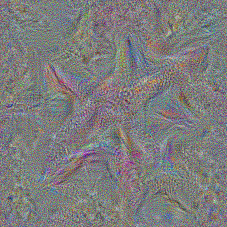} &
		\includegraphics[width=.2\textwidth]{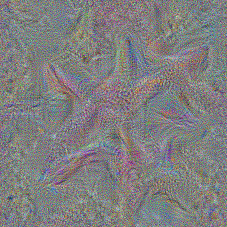} \\
		Iteration 0 & Iteration 10 & Iteration 50 & Iteration 100\\
	\end{tabular}
	\caption{The sequence of images sampled from the ``Starfish, sea star" category of the ``AlexNet" network \citep{krizhevsky2012imagenet} discriminatively trained on ImageNet ILSVRC-2012.}
	\label{fig:visualization_imagenet_iterations}
\end{figure}

Recently, researchers are interested in understanding what the machine learns. Suppose we care about the node at the top layer (the idea can be applied to the nodes at any layer). We consider generating samples from $p_y(x; w)$ with $w$ already learned by discriminative training (or any other methods). For this purpose, we need to assume a parametric reference distribution $q(x)$, such as Gaussian white noise distribution.
After discriminatively learning $f_y(x; w)$ for all $y$, we  can sample from the corresponding $p_y(x; w)$ by Hamiltonian Monte Carlo (HMC) \citep{neal2011mcmc}. 

Specifically, for any category $y$, we can write $p_y(x;w) \propto \exp(-U(x))$, where $U(x)=-f_y(x; w) +  |x|^2/(2 \sigma^2)$ ($\sigma$ is the standard deviation of $q(x)$). In physics context, $x$ is a position vector and $U(x)$ is the potential energy function. To implement Hamiltonian dynamics, we need to introduce an auxiliary momentum vector ${\phi}$ and the corresponding kinetic energy function $K({\phi})=|{\phi}|^2/2m$, where $m$ denotes the mass. Thus, a fictitious physical system described by the canonical coordinates $(x,{\phi})$ is defined, and its total energy is $H(x,{\phi})=U(x)+K({\phi})$. Each iteration of HMC draws a random sample from the marginal Gaussian distribution of ${\phi}$, and then evolve according to the Hamiltonian dynamics that conserves the total energy.

A key step in the leapfrog algorithm is the computation of the derivative of the potential energy function $\partial U/\partial x$, which includes calculating $\partial f_y(x; w)/\partial x$. The computation of $\partial f_y(x; w)/\partial x$ involves bottom-up convolution and max-pooling, followed by top-down deconvolution and arg-max un-pooling. The max-pooling and arg-max un-pooling are applied to the current synthesized image (not the input image, which is not needed by our method). The top-down derivative computation is derived from HMC, and is different from \citet{zeiler2013visualizing}. The visualization sequence of a category is shown in Fig. \ref{fig:visualization_imagenet_iterations}. 

\section{Experiments}

\subsection{Generative pre-training}
\label{sec:Generative pre-training}

In generative pre-training experiments, three different training approaches are studied: i) discriminative gradient ({\em DG}); ii) generative gradient ({\em GG}); iii) generative gradient pre-training + discriminative gradient refining ({\em GG+DG}). We build algorithms on the code of Caffe \citep{jia2014caffe} and the experiment settings are identical to \citet{jia2014caffe}. Experiments are performed on two commonly used image classification benchmarks:  MNIST \citep{lecun1998gradient} handwritten digit recognition and ImageNet ILSVRC-2012 \citep{deng2009imagenet} natural image classification.

\textbf{MNIST handwritten digit recognition.} We first study  generative pre-training on the MNIST dataset. The ``LeNet" network \citep{lecun1998gradient} is utilized, which is default for MNIST in Caffe. Although higher accuracy can be achieved by utilizing deeper networks, random image distortion etc, here we stick to the baseline network for fair comparison and experimental efficiency. Network training and testing are performed on the {\em train} and {\em test} sets respectively. For all the three training approaches, stochastic gradient descent is performed in training with a batch size of 64, a base learning rate of 0.01, a weight decay term of 0.0005, a momentum term of 0.9, and a max epoch number of 25. For {\em GG+DG}, the pre-training stage stops after 16 epochs and the discriminative gradient tuning stage starts with a base learning rate of 0.003.

The experimental results are presented in Table \ref{tab:err_MNIST}. The error rate of LeNet trained by discriminative gradient is 1.03\%. When trained by generative gradient, the error rate reduces to 0.85\%. When generative gradient pre-training and discriminative gradient refining are both applied, the error rate further reduces to 0.78\%, which is 0.25\% (24\% relatively) lower than that of discriminative gradient.

\begin{table}
\caption{Error rates on the MNIST {\em test} set of different training approaches utilizing the ``LeNet" network \citep{lecun1998gradient}.}
\label{tab:err_MNIST}
\renewcommand{\arraystretch}{1.2}
\begin{center}
\begin{tabular}{|p{3cm}|p{2cm}|p{2cm}|p{2cm}|}
\hline
Training approaches  &{\em DG} &{\em GG} & {\em GG+DG}\\
\hline
Error rates &1.03 &0.85 &\textbf{0.78}\\
\hline
\end{tabular}
\end{center}
\end{table}

\textbf{ImageNet ILSVRC-2012 natural image classification.} In experiments on ImageNet ILSVRC-2012, two networks are utilized, namely ``AlexNet" \citep{krizhevsky2012imagenet} and ``ZeilerFergusNet" (fast) \citep{zeiler2013visualizing}. Network training and testing are performed on the {\em train} and {\em val} sets respectively. In training, a single network is trained by stochastic gradient descent with a batch size of 256, a base learning rate of 0.01, a weight decay term of 0.0005, a momentum term of 0.9, and a max epoch number of 70. For {\em GG+DG}, the pre-training stage stops after 45 epochs and the discriminative gradient tuning stage starts with a base learning rate of 0.003. In testing, top-1 classification error rates are reported on the {\em val} set by classifying the center and the four corner crops of the input images.

\begin{table}
\renewcommand{\arraystretch}{1.2}
\newcommand{\tabincell}[2]{\begin{tabular}{@{}#1@{}}#2\end{tabular}}
\caption{Top-1 classification error rates on the ImageNet ILSVRC-2012 {\em val} set of different training approaches.}
\label{tab:err_ImageNet}
\begin{center}
\begin{tabular}{|p{3.5cm}|p{2cm}|p{2cm}|p{2cm}|}
\hline
Training approaches  &{\em DG} &{\em GG} & {\em GG+DG}\\
\hline
AlexNet &40.7 &45.8 & \textbf{39.6}\\
\hline
ZeilerFergusNet (fast) &38.4 &44.3 & \textbf{37.4}\\
\hline
\end{tabular}
\end{center}
\end{table}

As shown in Table \ref{tab:err_ImageNet}, the error rates of discriminative gradient training applied on AlexNet and ZeilerFergusNet are 40.7\% and 38.4\% respectively, while the error rates of generative gradient are 45.8\% and 44.3\% respectively. Generative gradient pre-training followed by discriminative gradient refining achieves error rates of 39.6\% and 37.4\% respectively, which are 1.1\% and 1.0\% lower than those of discriminative gradient.

Experiment results on MNIST and ImageNet ILSVRC-2012 show that generative gradient pre-training followed by discriminative gradient refining  improves the classification accuracies for varying networks. At the beginning stage of training, updating network parameters according to the generative gradient provides useful pre-training, which leads the network parameters toward a good local optimum. 

As to the computational cost, generative gradient is on par with discriminative gradient. The computational cost of the generative loss layer itself is ignorable in the network compared to the computation at the convolutional layers and the fully-connected layers. The total epoch numbers of {\em GG+DG} is on par with that of {\em DG}. 

\begin{figure}
	\centering
	\includegraphics[width=.043\textwidth]{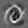}
	\includegraphics[width=.043\textwidth]{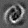}
	\includegraphics[width=.043\textwidth]{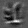}
	\includegraphics[width=.043\textwidth]{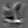}
	\includegraphics[width=.043\textwidth]{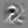}
	\includegraphics[width=.043\textwidth]{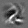}
	\includegraphics[width=.043\textwidth]{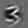}
	\includegraphics[width=.043\textwidth]{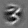}
	\includegraphics[width=.043\textwidth]{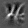}
	\includegraphics[width=.043\textwidth]{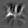}
	\includegraphics[width=.043\textwidth]{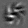}
	\includegraphics[width=.043\textwidth]{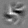}
	\includegraphics[width=.043\textwidth]{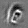}
	\includegraphics[width=.043\textwidth]{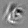}
	\includegraphics[width=.043\textwidth]{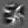}
	\includegraphics[width=.043\textwidth]{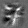}
	\includegraphics[width=.043\textwidth]{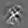}
	\includegraphics[width=.043\textwidth]{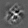}
	\includegraphics[width=.043\textwidth]{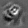}
	\includegraphics[width=.043\textwidth]{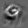}
	\caption{Samples from the nodes at the final fully-connected layer in the fully trained LeNet model, which correspond to different handwritten digits.}
	\label{fig:visualization_mnist_category}
\end{figure}

\begin{figure}
\centering
\setlength{\fboxrule}{1pt}
\setlength{\fboxsep}{0cm}
\subfloat[conv1]{
\begin{minipage}{.35\textwidth}
\fbox{\begin{minipage}{.22\textwidth}
\includegraphics[height=.28\textwidth]{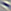}
\includegraphics[height=.28\textwidth]{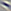}
\includegraphics[height=.28\textwidth]{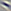}
\end{minipage}}
\fbox{\begin{minipage}{.22\textwidth}
\includegraphics[height=.28\textwidth]{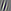}
\includegraphics[height=.28\textwidth]{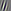}
\includegraphics[height=.28\textwidth]{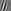}
\end{minipage}}
\fbox{\begin{minipage}{.22\textwidth}
\includegraphics[height=.28\textwidth]{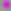}
\includegraphics[height=.28\textwidth]{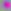}
\includegraphics[height=.28\textwidth]{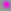}
\end{minipage}}
\fbox{\begin{minipage}{.22\textwidth}
\includegraphics[height=.28\textwidth]{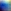}
\includegraphics[height=.28\textwidth]{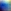}
\includegraphics[height=.28\textwidth]{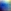}
\end{minipage}}
\end{minipage}}
\subfloat[conv2]{
\begin{minipage}{.6\textwidth}
\fbox{\begin{minipage}{.22\textwidth}
\includegraphics[height=.28\textwidth]{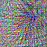}
\includegraphics[height=.28\textwidth]{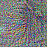}
\includegraphics[height=.28\textwidth]{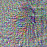}
\end{minipage}}
\fbox{\begin{minipage}{.22\textwidth}
\includegraphics[height=.28\textwidth]{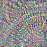}
\includegraphics[height=.28\textwidth]{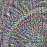}
\includegraphics[height=.28\textwidth]{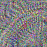}
\end{minipage}}
\fbox{\begin{minipage}{.22\textwidth}
\includegraphics[height=.28\textwidth]{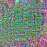}
\includegraphics[height=.28\textwidth]{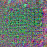}
\includegraphics[height=.28\textwidth]{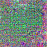}
\end{minipage}}
\fbox{\begin{minipage}{.22\textwidth}
	\includegraphics[height=.28\textwidth]{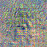}
	\includegraphics[height=.28\textwidth]{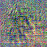}
	\includegraphics[height=.28\textwidth]{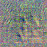}
\end{minipage}}
\end{minipage}}\\
\subfloat[conv3]{
\begin{minipage}{.5\textwidth}
\fbox{\begin{minipage}{.3\textwidth}
\includegraphics[height=.30\textwidth]{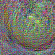}
\includegraphics[height=.30\textwidth]{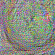}
\includegraphics[height=.30\textwidth]{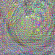}
\end{minipage}}
\fbox{\begin{minipage}{.3\textwidth}
	\includegraphics[height=.30\textwidth]{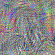}
	\includegraphics[height=.30\textwidth]{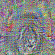}
	\includegraphics[height=.30\textwidth]{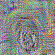}
\end{minipage}}
\fbox{\begin{minipage}{.3\textwidth}
		\includegraphics[height=.30\textwidth]{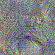}
		\includegraphics[height=.30\textwidth]{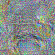}
		\includegraphics[height=.30\textwidth]{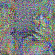}
	\end{minipage}}
\end{minipage}}
\subfloat[conv4]{
	\begin{minipage}{.5\textwidth}
		\fbox{\begin{minipage}{.3\textwidth}
				\includegraphics[height=.30\textwidth]{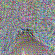}
				\includegraphics[height=.30\textwidth]{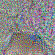}
				\includegraphics[height=.30\textwidth]{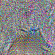}
			\end{minipage}}
			\fbox{\begin{minipage}{.3\textwidth}
					\includegraphics[height=.30\textwidth]{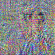}
					\includegraphics[height=.30\textwidth]{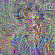}
					\includegraphics[height=.30\textwidth]{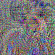}
				\end{minipage}}
				\fbox{\begin{minipage}{.3\textwidth}
						\includegraphics[height=.30\textwidth]{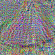}
						\includegraphics[height=.30\textwidth]{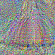}
						\includegraphics[height=.30\textwidth]{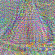}
					\end{minipage}}
				\end{minipage}}\\
				\subfloat[conv5]{
					\begin{minipage}{1.0\textwidth}
						\fbox{\begin{minipage}{.23\textwidth}
								\includegraphics[height=.30\textwidth]{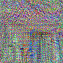}
								\includegraphics[height=.30\textwidth]{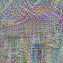}
								\includegraphics[height=.30\textwidth]{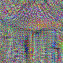}
							\end{minipage}}
							\fbox{\begin{minipage}{.23\textwidth}
									\includegraphics[height=.30\textwidth]{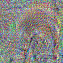}
									\includegraphics[height=.30\textwidth]{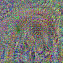}
									\includegraphics[height=.30\textwidth]{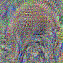}
								\end{minipage}}
								\fbox{\begin{minipage}{.23\textwidth}
										\includegraphics[height=.30\textwidth]{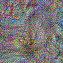}
										\includegraphics[height=.30\textwidth]{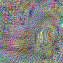}
										\includegraphics[height=.30\textwidth]{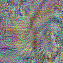}
									\end{minipage}}
									\fbox{\begin{minipage}{.23\textwidth}
											\includegraphics[height=.30\textwidth]{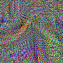}
											\includegraphics[height=.30\textwidth]{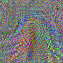}
											\includegraphics[height=.30\textwidth]{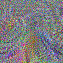}
										\end{minipage}}
									\end{minipage}}
									\caption{Samples from the nodes at the intermediate convolutional layers (conv1 to conv5) in the fully trained AlexNet model. }
									\label{fig:visualization_imagenet_intermediate}
								\end{figure}

\begin{figure}
	\centering
	\subfloat[Hen]{
		\includegraphics[width=.25\textwidth]{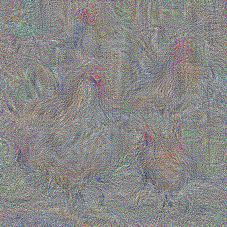}
		\includegraphics[width=.25\textwidth]{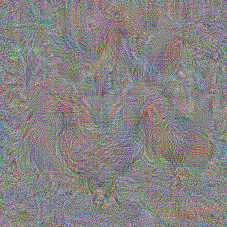}
		\includegraphics[width=.25\textwidth]{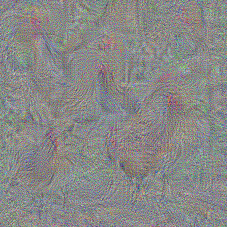}
		\includegraphics[width=.25\textwidth]{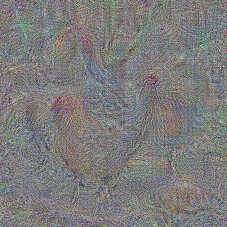}
	}\\
	\subfloat[Ostrich]{
		\includegraphics[width=.25\textwidth]{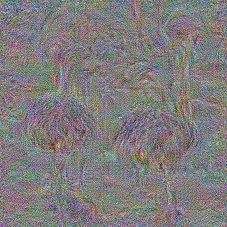}
		\includegraphics[width=.25\textwidth]{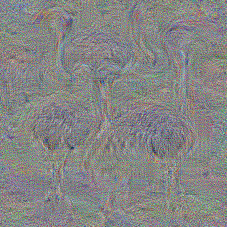}
		\includegraphics[width=.25\textwidth]{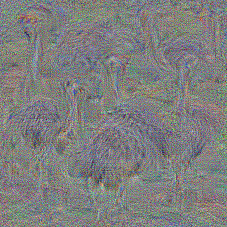}
		\includegraphics[width=.25\textwidth]{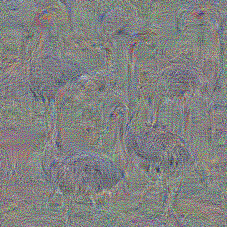}
	}\\
	\subfloat[Fish]{
		\includegraphics[width=.25\textwidth]{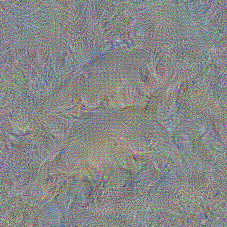}
		\includegraphics[width=.25\textwidth]{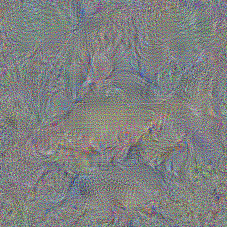}
		\includegraphics[width=.25\textwidth]{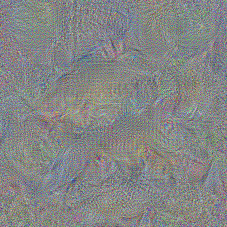}
		\includegraphics[width=.25\textwidth]{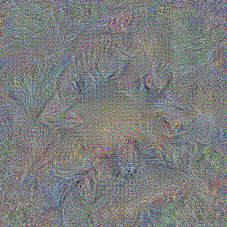}
	}\\
	\subfloat[Horse cart]{
		\includegraphics[width=.25\textwidth]{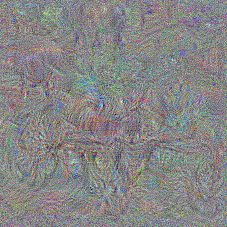}
		\includegraphics[width=.25\textwidth]{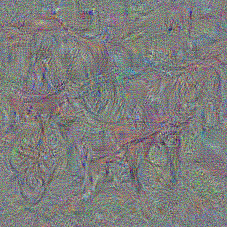}
		\includegraphics[width=.25\textwidth]{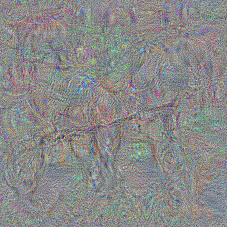}
		\includegraphics[width=.25\textwidth]{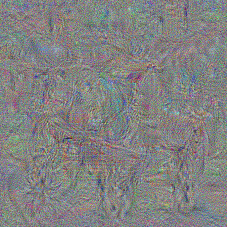}
	}\\
	\caption{Samples from the nodes at the final fully-connected layer (fc8) in the fully trained AlexNet model. {More examples are included in the supplementary materials.}}
	\label{fig:visualization_imagenet_category}
\end{figure}

\subsection{Generative visualization}

In the generative visualization experiments, we visualize the nodes of the LeNet network and the AlexNet network trained by discriminative gradient on MNIST and ImageNet ILSVRC-2012 respectively. The algorithm can visualize networks trained by generative gradient as well.

We first visualize the nodes at the final fully-connected layer of LeNet. In the experiments, we delete the drop-out layer to avoid unnecessary noise for visualization. At the beginning of visualization, $x$ is initialized by Gaussian distribution with standard deviation 10. The HMC iteration number, the leapfrog step size, the leapfrog step number, the standard deviation of the reference distribution $\sigma$, and the particle mass are set to be 300, 0.0001, 100, 10, and 0.0001 respectively. The visualization results are shown in Fig. \ref{fig:visualization_mnist_category}. 

We further visualize the nodes in AlexNet, which is a much larger network compared to LeNet. Both nodes from the intermediate convolutional layers (conv1 to conv5) and the final fully-connected layer (fc8) are visualized. To visualize the intermediate layers, for instance the layer conv2 with 256 filters, all layers above conv2 are removed other than the generative visualization layer. The size of the synthesized images are  designed so that the dimension of the response from conv2 is $1 \times 1 \times 256$. We can visualize each filter by assigning label from 1 to 256. The leapfrog step size, the leapfrog step number, the standard deviation of the reference distribution $\sigma$, and the particle mass are set to be 0.000003, 50, 10, and 0.00001 respectively. The HMC iteration numbers are 100 and 500 for nodes from the intermediate convolutional and the final fully-connected layer respectively. The synthesized images for the final layer are initialized from the zero image.

The samples from the intermediate convolutional layers and the final fully-connected layer of AlexNet are shown in Fig. \ref{fig:visualization_imagenet_intermediate} and \ref{fig:visualization_imagenet_category} respectively. The HMC algorithm produces meaningful and varied samples, which reveals what is learned by the nodes at different layers of the network. Note that such samples are generated from the trained model directly, without using a large hold-out collection of images as in \citet{girshick2014rich,zeiler2013visualizing,long2014convnets}.

As to the computational cost, it varies for nodes at different layers within different networks. On a desktop with GTX Titian, it takes about 0.4 minute to draw a sample for nodes at the final fully-connected layer of LeNet. In AlexNet, for nodes at the first convolutional layer and at the final fully-connected layer, it takes about 0.5 minute and 12 minute to draw a sample respectively. The code can be downloaded at \url{http://www.stat.ucla.edu/~yang.lu/Project/generativeCNN/main.html}

\section{Conclusion}

Given the recent successes of CNNs, it is worthwhile to explore their generative aspects. In this work, we show that a simple generative model can be constructed based on the CNN. The generative model helps to pre-train the CNN. It also helps to visualize the knowledge of the learned CNN.

The proposed visualizing scheme can sample from the generative model, and it may be turned into a parametric generative learning algorithm, where the generative gradient can be approximated by samples generated by the current model. 

\section*{Acknowledgement}

The work is supported by NSF DMS 1310391, ONR MURI N00014-10-1-0933, DARPA MSEE FA8650-11-1-7149.

\bibliography{iclr2015}
\bibliographystyle{iclr2015}

\newpage
\section*{Supplementary Materials}

\setcounter{figure}{0}    
\renewcommand{\thefigure}{B\arabic{figure}}

\subsection*{A. Discriminative vs generative log-likelihood and gradient for batch training}

During training, on a batch of training examples, $\{(x_i, y_i), i = 1, ..., n\}$, the generative log-likelihood is
\[
l_{\G}(w) = \sum_i \log p(x_i|y_i, w) = \sum_i \log \frac{ \exp\left(f_{y_i}(x_i; w)\right) }{Z_{y_i}(w)} \approx \sum_i  \log \frac{ \exp\left(f_{y_i}(x_i; w)\right) }{\sum_{k}  \exp\left(f_{y_i}(x_k; w)\right)/n}.
\]
The gradient with respect to $w$ is
\[
l'_{\G}(w) = \sum_i  \Bigg[\frac{\partial}{\partial w}  f_{y_i}(x_i; w)  - \sum_j \frac{\partial}{\partial w}  f_{y_i}({x}_j; w)  \frac{\exp(f_{y_i}({x}_j; w))}{\sum_k \exp(f_{y_i}({x}_k; w))}\Bigg].
\]

The discriminative log-likelihood is
\[
l_{\D}(w) = \sum_i \log p(y_i|x_i, w) = \sum_i \log \frac{\exp(f_{y_i}(x_i; w) )}{\sum_y \exp(f_y(x_i; w) )}.
\]
The gradient with respect to $w$ is
\[
l'_{\D}(w) = \sum_i  \Bigg[ \frac{\partial}{\partial w} f_{y_i}(x_i; w) - \sum_{y}  \frac{\partial}{\partial w} f_y(x_i; w)    \frac{\exp(f_y(x_i; w))}{\sum_{y} \exp(f_y(x_i; w))} \Bigg].
\]
$l'_{\D}$ and $l'_{\G}$ are similar in form but different in the summation operations. In $l'_{\D}$, the summation is over category $y$ while $x_i$ is fixed, whereas in $l'_{\G}$, the summation is over example $x_j$ while $y_i$ is fixed.

In the generative gradient, we want $f_{y_i}$ to assign high score to $x_i$ as well as those observations that belong to $y_i$, but assign low scores to those observations that do not belong to $y_i$. This constraint is for the {same} $f_{y_i}$, regardless of what other $f_y$ do for $y \neq y_i$.

In the discriminative gradient, we want $f_y(x_i)$ to work together for all {different} $y$, so that $f_{y_i}$ assigns high score to $x_i$ than other $f_y$ for $y \neq y_i$.

Apparently, the discriminative constraint is weaker because it involves all $f_y$, and the generative constraint is stronger because it involves single $f_y$. After generative learning, these $f_y$ are well behaved and then we can continue to refine them (including the intercepts for different $y$) to satisfy the discriminative constraint.

\subsection*{B. More generative visualization examples}

\begin{figure}
	\centering
	\subfloat[Boat]{
		\includegraphics[width=.25\textwidth]{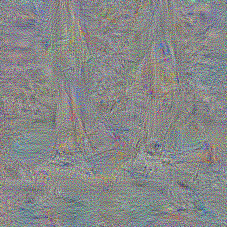}
		\includegraphics[width=.25\textwidth]{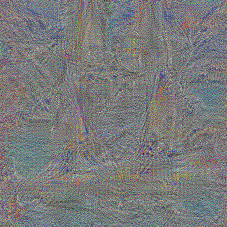}
		\includegraphics[width=.25\textwidth]{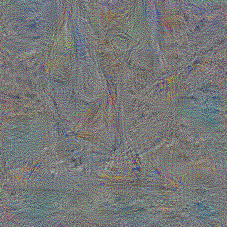}
		\includegraphics[width=.25\textwidth]{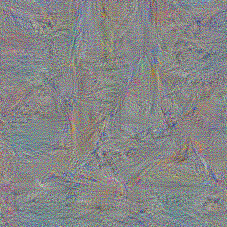}
	}\\
	\subfloat[Peacock]{
		\includegraphics[width=.25\textwidth]{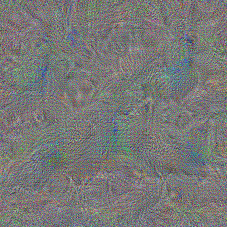}
		\includegraphics[width=.25\textwidth]{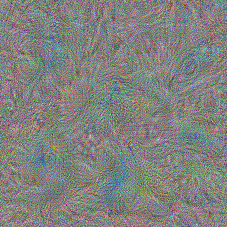}
		\includegraphics[width=.25\textwidth]{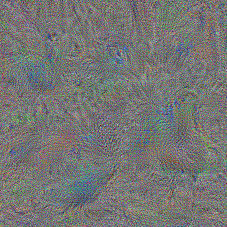}
		\includegraphics[width=.25\textwidth]{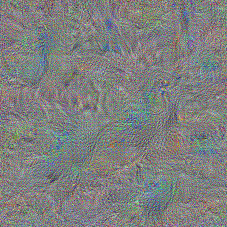}
	}\\
	\subfloat[Panda]{
		\includegraphics[width=.25\textwidth]{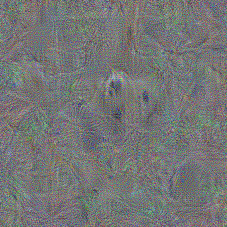}
		\includegraphics[width=.25\textwidth]{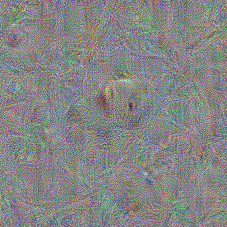}
		\includegraphics[width=.25\textwidth]{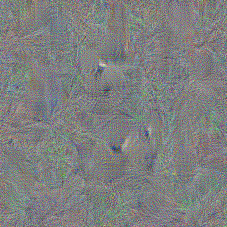}
		\includegraphics[width=.25\textwidth]{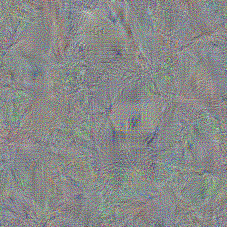}
	}\\
	\subfloat[Orange]{
		\includegraphics[width=.25\textwidth]{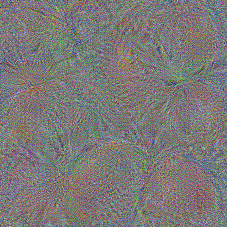}
		\includegraphics[width=.25\textwidth]{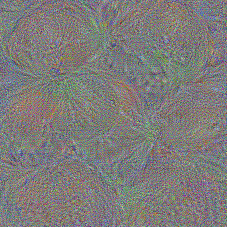}
		\includegraphics[width=.25\textwidth]{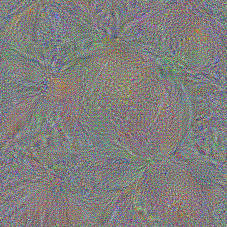}
		\includegraphics[width=.25\textwidth]{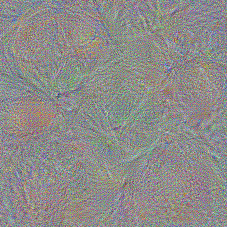}
	}\\
	\caption{More samples from the nodes at the final fully-connected layer (fc8) in the fully trained AlexNet model, which correspond to different object categories (part 1).}
	\label{fig:visualization_imagenet_category_1}
\end{figure}

\begin{figure}
	\centering
	\subfloat[Lotion bottle]{
		\includegraphics[width=.25\textwidth]{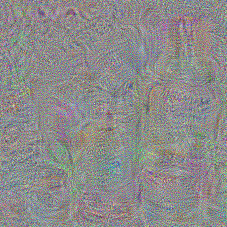}
		\includegraphics[width=.25\textwidth]{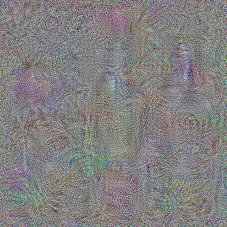}
		\includegraphics[width=.25\textwidth]{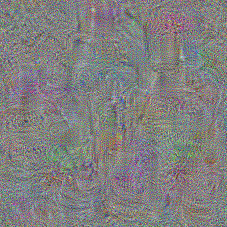}
		\includegraphics[width=.25\textwidth]{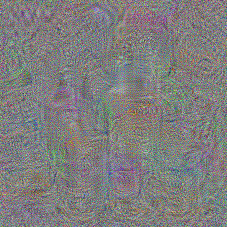}
	}\\
	\subfloat[Hook]{
		\includegraphics[width=.25\textwidth]{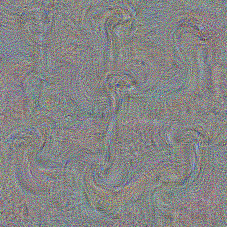}
		\includegraphics[width=.25\textwidth]{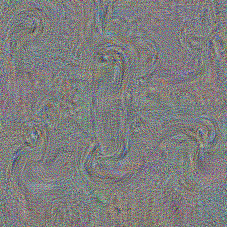}
		\includegraphics[width=.25\textwidth]{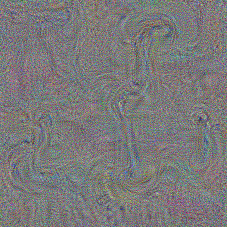}
		\includegraphics[width=.25\textwidth]{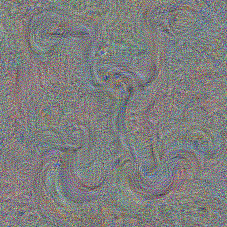}
	}\\
	\subfloat[Lawn mower]{
		\includegraphics[width=.25\textwidth]{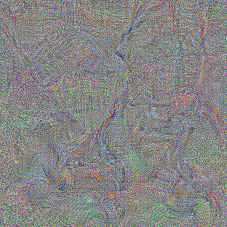}
		\includegraphics[width=.25\textwidth]{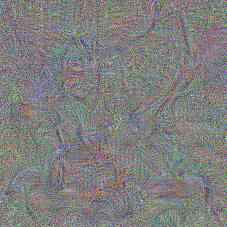}
		\includegraphics[width=.25\textwidth]{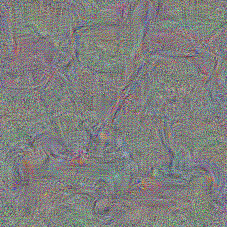}
		\includegraphics[width=.25\textwidth]{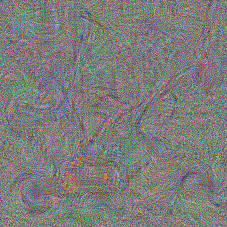}
	}\\
	\subfloat[Hourglass]{
		\includegraphics[width=.25\textwidth]{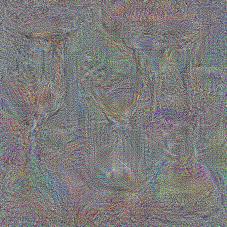}
		\includegraphics[width=.25\textwidth]{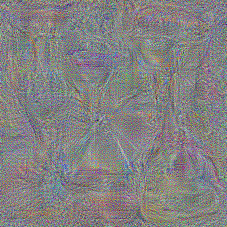}
		\includegraphics[width=.25\textwidth]{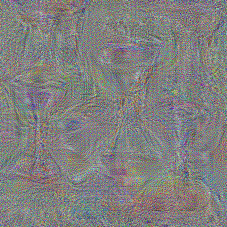}
		\includegraphics[width=.25\textwidth]{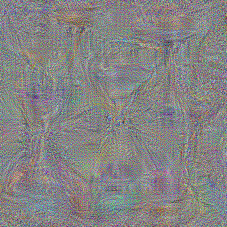}
	}\\	
	\caption{More samples from the nodes at the final fully-connected layer (fc8) in the fully trained AlexNet model, which correspond to different object categories (part 2).}
	\label{fig:visualization_imagenet_category_2}
\end{figure}

\begin{figure}
	\centering
	\subfloat[Knot]{
		\includegraphics[width=.25\textwidth]{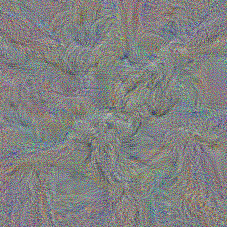}
		\includegraphics[width=.25\textwidth]{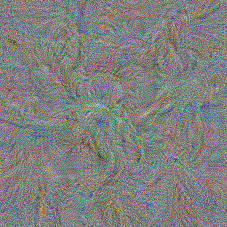}
		\includegraphics[width=.25\textwidth]{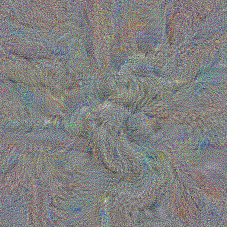}
		\includegraphics[width=.25\textwidth]{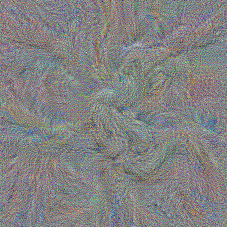}
	}\\
	\subfloat[Nail]{
		\includegraphics[width=.25\textwidth]{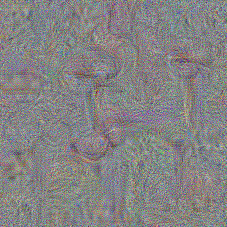}
		\includegraphics[width=.25\textwidth]{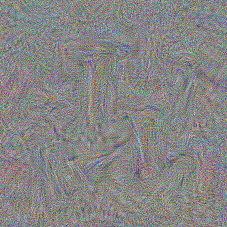}
		\includegraphics[width=.25\textwidth]{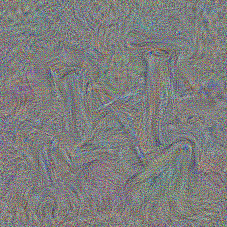}
		\includegraphics[width=.25\textwidth]{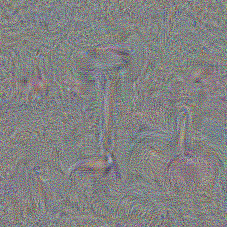}
	}\\
	\subfloat[Academic gown]{
		\includegraphics[width=.25\textwidth]{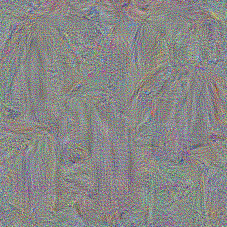}
		\includegraphics[width=.25\textwidth]{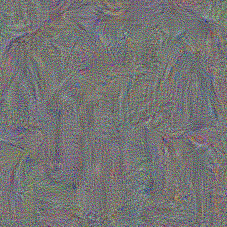}
		\includegraphics[width=.25\textwidth]{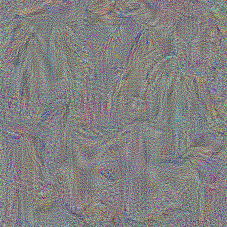}
		\includegraphics[width=.25\textwidth]{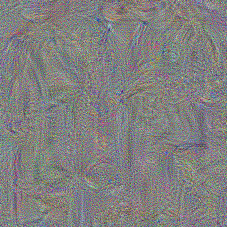}
	}\\
	\subfloat[Goose]{
		\includegraphics[width=.25\textwidth]{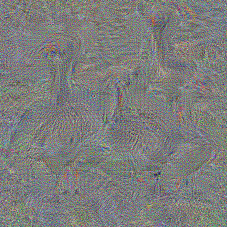}
		\includegraphics[width=.25\textwidth]{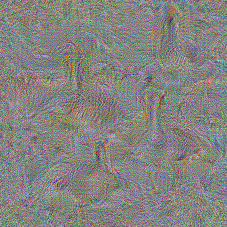}
		\includegraphics[width=.25\textwidth]{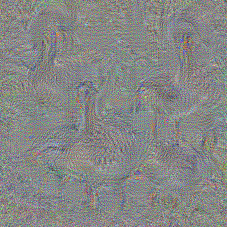}
		\includegraphics[width=.25\textwidth]{./figures/Goose/caffe_imagenet_visualization_display_99_iter_299_step_51_sample_9.png}
	}\\
	\caption{More samples from the nodes at the final fully-connected layer (fc8) in the fully trained AlexNet model, which correspond to different object categories (part 3).}
	\label{fig:visualization_imagenet_category_3}
\end{figure}

More generative visualization examples for the nodes at the final fully-connected layer in the fully trained AlexNet model are shown in Fig. \ref{fig:visualization_imagenet_category_1}, Fig. \ref{fig:visualization_imagenet_category_2} and Fig. \ref{fig:visualization_imagenet_category_3}.

\end{document}